\documentclass{article}

\usepackage{microtype}
\usepackage{graphicx}
\usepackage{subfigure}
\usepackage{booktabs} 
\usepackage{xcolor}
\usepackage[hyphens]{url}

\usepackage{hyperref}

\newcommand{\PT}[1]{{\color{magenta}\bfseries [Po-An:: #1]}} 

\usepackage[accepted]{mlsys2024}

\usepackage{comment}

\mlsystitlerunning{SDQ: Sparse Decomposed Quantization for LLM Inference}

\begin{document}

\twocolumn[
\mlsystitle{SDQ: Sparse Decomposed Quantization for LLM Inference}



\mlsyssetsymbol{equal}{*}

\begin{mlsysauthorlist}
\mlsysauthor{Geonhwa Jeong}{gt}
\mlsysauthor{Po-An Tsai}{nv}
\mlsysauthor{Stephen W. Keckler}{nv}
\mlsysauthor{Tushar Krishna}{gt}
\end{mlsysauthorlist}

\mlsysaffiliation{gt}{School of Computer Science, Georgia Institute of Technology, United States}
\mlsysaffiliation{nv}{Architecture Research Group, NVIDIA, United States}

\mlsyscorrespondingauthor{Geonhwa Jeong}{geonhwa.jeong@gatech.edu}

\mlsyskeywords{Machine Learning, MLSys}

\vskip 0.3in

\begin{abstract}
Recently, large language models (LLMs) have shown surprising performance in task-specific workloads as well as general tasks with the given prompts. 
However, to achieve unprecedented performance, recent LLMs use billions to trillions of parameters, which hinder the wide adaptation of those models due to their extremely large compute and memory requirements.
To resolve the issue, various model compression methods are being actively investigated.
In this work, we propose \textbf{SDQ} (\textbf{S}parse \textbf{D}ecomposed \textbf{Q}uantization) to exploit both structured sparsity and quantization to achieve both high compute and memory efficiency.
From our evaluations, we observe that SDQ can achieve 4$\times$ effective compute throughput with $<$1\% quality drop.

\end{abstract}
]



\printAffiliationsAndNotice{}  

\section{Introduction}
\label{sec:intro}
Large Language Models (LLMs)~\cite{brown2020language,chowdhery2022palm,touvron2023llama2} with billions or trillions of parameters have gained extensive attention as they show promising quality in various domains.
With such popularity, efficiently deploying and accelerating LLMs are the utmost research question for system designers, as the large number of weights causes a huge memory footprint and an enormous amount of computations.
To address this issue, recent work has applied various model compression methods, such as sparsification~\cite{frantar2023sparsegpt} and quantization~\cite{llmint8_nipes22}, on LLMs to reduce memory footprint and computations.
However, when compared to classic DNN models (ResNet, BERT), LLMs introduce new challenges to both sparsification and quantization.

Model sparsification, such as model pruning~\cite{NIPS1989_obd, NIPS2015_songhan}, reduces the target model size by removing parameters in weights and storing the weights using a compressed format.
Sparsity not only reduces the memory requirement but also enables performance improvement and power saving by skipping ineffectual computation, i.e., $0 \times X = 0$.
Exploiting this opportunity, recent NVIDIA Ampere GPU systems provide native support for 2:4 structured sparsity, a specialized form of sparse pattern that guarantees $2\times$ computation reduction and performance gain with low overhead, i.e., sparsity tax~\cite{wu2023highlight}.

\begin{figure}[!t]
	\centering
	\includegraphics[width=0.49\textwidth]{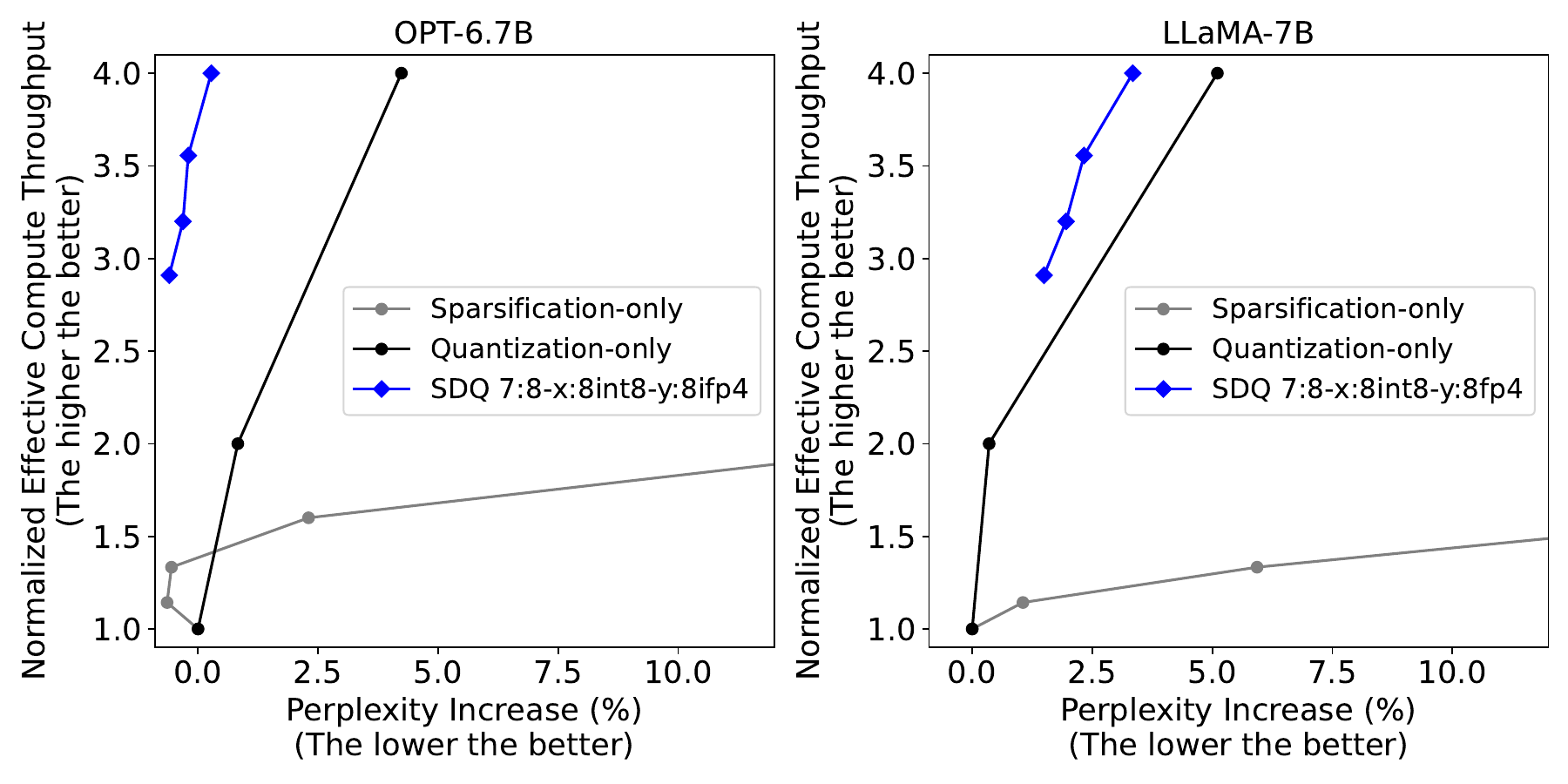}
\vspace{-1em}
        \caption{Effective compute throughput and perplexity increase comparison of sparsification-only, quantization-only, and SDQ on OPT-6.7B and LLaMA-7B.}
\vspace{-1em}
\label{fig:sdq_highlight}
\end{figure}

Although many sparsification methods have been proposed for the classic DNNs, these techniques are less successful so far to compress LLMs. Previous efforts~\cite{Hoefler} have shown how to compress classic DNNs by more than 90\% (10$\times$ computation reduction) with limited loss of model quality; however, when it comes to LLMs, compressing beyond 50\% (2$\times$ computation reduction) with a limited loss of accuracy is already challenging~\cite{frantar2023sparsegpt,sun2023wanda} as shown in \autoref{fig:sdq_highlight}.


On the other hand, quantization reduces the target model size by using a narrower bit width per value. For example, if we use an 8-bit or a 4-bit format instead of a 16-bit format for each parameter, we can reduce the model size by half and a quarter, respectively.
Furthermore, when quantization is applied to both weights and activations, we can use specialized, low-bit-width arithmetic computation instead of full-precision computation, which is both more power-efficient and area-efficient~\cite{Horowitz, vanbaalen2023fp8}.
As the low-bit-width arithmetic unit takes a smaller area, processors with a given area budget for compute units could allocate more low-bit-width compute units.
For example, in NVIDIA Ampere GPU, the peak tensor throughput for \texttt{int8} format is 2$\times$ of the peak throughput for \texttt{fp16} (4$\times$ for \texttt{int4} format).
 
Like sparsification, quantization also faces new challenges with LLMs. While prior work has shown how to use various 4-bit formats~\cite{dai2021vsq,ms2023isca} for classic models, limited work shows how to quantize \emph{both} activation and weights of LLMs to 4-bit format to leverage the high throughput, low-bit-width hardware units.
Recent work~\cite{guo2023olive, lin2023awq} has shown that \emph{outliers} in activations obstruct quantizing both activations and weights as it causes high quantization error, which results in a huge error in the final output when propagated.
Various work in this area instead focuses on 8-bit dual quantization~\cite{llmint8_nipes22}, or quantization for weights only~\cite{lin2023awq, dettmers2023spqr},
limiting the computation reduction to be up to 2$\times$.
\textbf{In this paper}, we demonstrate how to achieve even larger computation reduction (more than 2$\times$) by combining sparsification and quantization. 
Our key insight is that sparsification could be used to complement the outlier problem in the LLM quantization while maintaining high throughput with a low overhead using the structured sparsity patterns. 
Based on the insight, we propose a hybrid model compression method leveraging both sparsification and quantization to further increase computation reduction, while maintaining model quality through decomposing weights into structured sparse tensors and using structured sparse/low-bit-width compute HW.
We focus on post-training model compression, which does not require any model retraining or fine-tuning as training LLMs could be infeasible in many cases (due to limited compute resources or data).

We summarize our contribution in the following:
\begin{itemize}
    \item We illustrate the opportunity to treat the outliers during quantization as structured sparse tensors to accelerate with structured sparse HW.
    \item We propose SDQ, a technique to combine both sparse and quantization emerging hardware support. It achieves a better Pareto curve in model quality and compute throughput than sparsification-only or quantization-only methods.
    \item We show that SDQ is orthogonal to sparsification and quantization techniques. With an improved sparsified or quantized model, SDQ would also perform better.
\end{itemize}

\section{Background}
\begin{figure}[!t]
	\centering
	\includegraphics[width=0.49\textwidth]{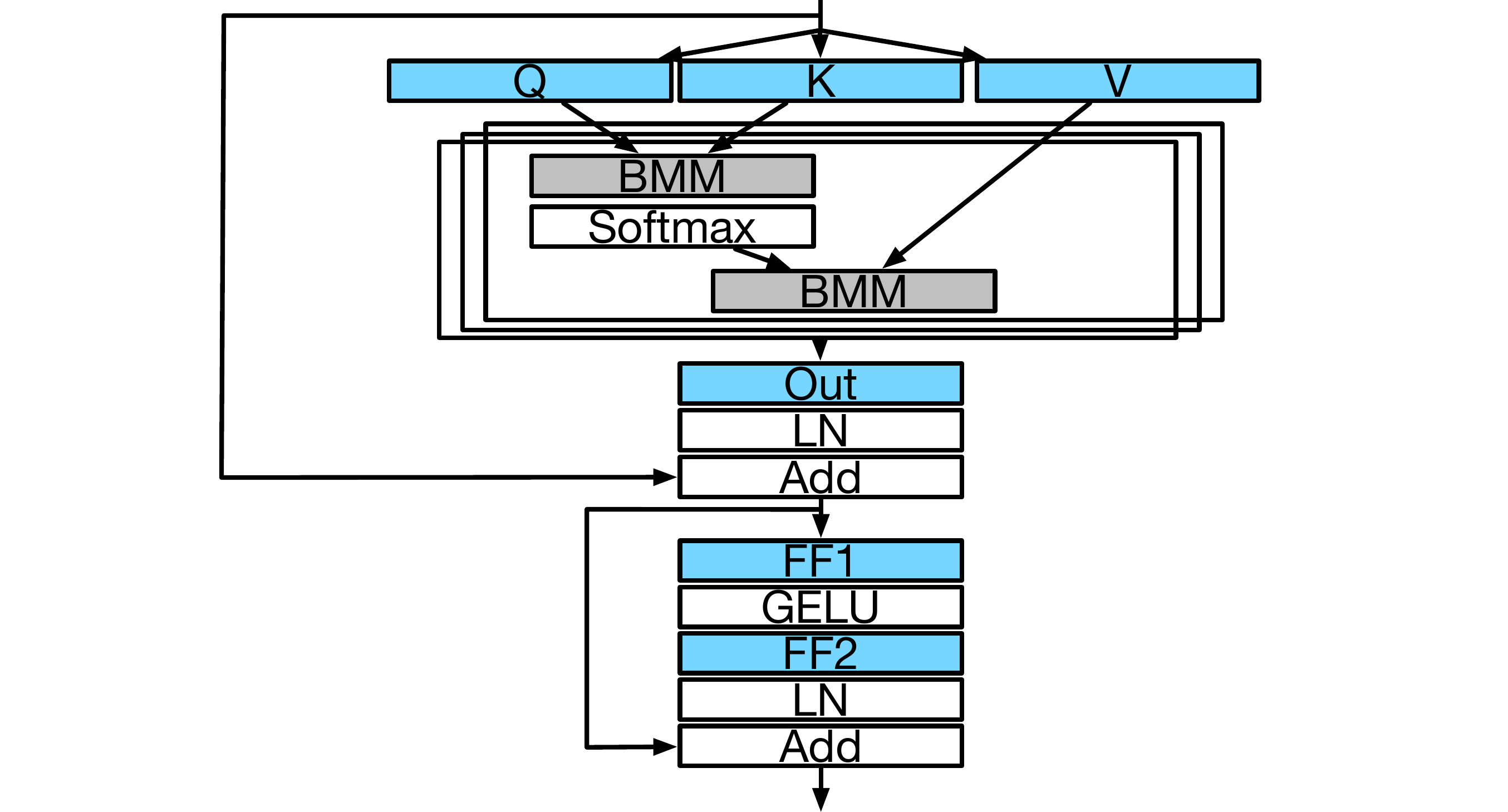}
\vspace{-1em}
        \caption{A transformer block architecture.}
    \label{fig:transformer}
\vspace{-1em}

\end{figure}
\label{sec:background}
\subsection{Compressing LLM models}
In \autoref{fig:transformer}, we show a high-level overview of decoder-only Transformer block architecture used in the recent LLMs.
Each Transformer block contains multiple layers including linear layers, activation layers, normalization layers, etc., and 
the layers colored with blue and grey in \autoref{fig:transformer} contribute to $>$99\% of the computations during inference. 
Linear layers ({\color{blue}Q, K, V, out, FF1, and FF2}) consist of matrix-matrix multiplications (GEMMs) with static weights, while self-attention layers ({\color{gray} BMMs}) consists of GEMMs with dynamic activations. 
While the actual computation breakdown depends on the input context sizes, as well as the model parameters, generally speaking, GEMMs from the linear layers dominates the overall computation and latency~\cite{transformer_nips17}. As the result, in this paper, we consider model optimization and compression for only the GEMMs for the linear layers (blue layers with static weights).

\begin{table*}
    \centering
    \begin{tabular}{|c|c|c|c|c|c|c|} \hline 
    & Sparsification& Quantization & Outlier & Compute &Compute\\ 
    & Configuration & Configuration & Extraction & Cores&Bit Width\\ \hline 
         ASP~\cite{asp}&  2:4&  X& X & 2:4 Sparse TC&16b\\ \hline 
         WANDA~\cite{sun2023wanda}&  2:4&  X& X& 2:4 Sparse TC&16b\\ \hline 
         VS-Quant~\cite{dai2021vsq}&  X&  W4-8A4-8& X& TC&4b/8b\\ \hline 
         AWQ~\cite{lin2023awq}&  X&  W3-4A16& X& TC&16b\\ \hline
 SparseGPT~\cite{frantar2023sparsegpt}& 2:4/4:8& W4A16& X& 2:4 Sparse TC&16b\\\hline
 LLM.int8()~\cite{llmint8_nipes22}& Unstr. 0.1\%& W8A8& O& TC + CUDA Core&8b\\\hline
 SpQR~\cite{dettmers2023spqr}& Unstr. 1\%& W3-4A16& O& TC + CUDA Core&16b\\\hline
 OWQ~\cite{lee2023owq}& Col-wise $<$1\%& W3-4A16& O& TC + CUDA Core&16b\\\hline
 SqueezeLLM~\cite{kim2023squeezellm}& Unstr. $<$1\%& W3-4A16& O& TC + CUDA Core&16b\\\hline
 SDQ (Our Work)& N:4/N:8& W4A4/W8A8& O& Flexible Sparse TC&4b/8b\\\hline
    \end{tabular}
    \vspace{-0.5em}
    \caption{Summary of related work on sparsification and quantization. TC means Tensor Core and Unstr. means Unstructured sparsity.}
    \vspace{-0.5em}
    \label{tab:previous_work}
\end{table*}
\subsection{Sparification}
Sparsification induces sparsity in tensors, i.e. making some non-zero elements to zeros.
The most popular sparsification technique used for DNNs is weight pruning which statically removes non-zero weight values by setting them zeros. 
Pruning DNN models is effective for classic DNNs as they are often overly parameterized, so removing \textbf{insignificant values} would not impact model quality when done carefully.

\begin{figure}[!t]
	\centering
	\includegraphics[width=0.49\textwidth]{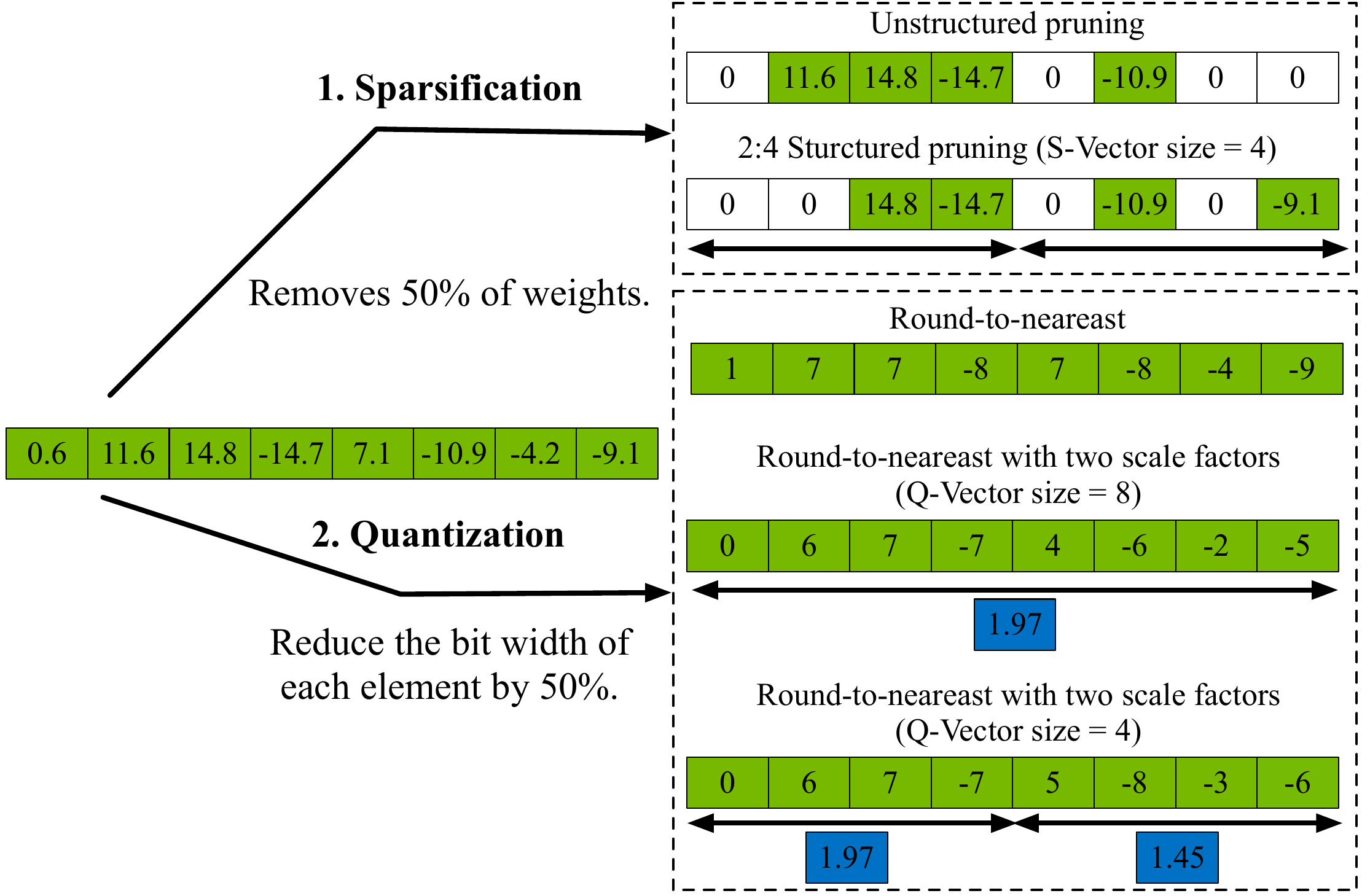}
\vspace{-1em}
        \caption{Overview of sparsification and quantization.}
\vspace{-1em}
    \label{fig:sparse-quant}
\end{figure}

The simplest way to perform pruning is by choosing insignificant values based on the magnitudes~\cite{NIPS2015_songhan} as shown in \autoref{fig:sparse-quant}, but previous work has proposed various metrics to decide how to choose values to be removed.
For example, some methods leverage input samples and use the first order error~\cite{molchanov2022lana} during inference to determine which weigh value can be pruned with minimal error.
Other method proposes to use second order error (Hessian)~\cite{frantar2023sparsegpt} to predict impact of weight pruning more accurately, but at a higher computation cost, since calculating the Hessian matrix is non-trivial.
While these pruning methods uses different metrics, they are very successful in compressing the conventional DNN models: many of them show possibility to compress more than 10$\times$.

Unlike a plethora of work on sparsifying conventional DNNs such as CNNs, there is a still limited number of work on sparisfying LLMs due to two major challeneges.
First, LLMs are usually less over-parameterized~\cite{chowdhery2022palm} than the classic models.
Therefore, most of the weight values are relatively crucial and cannot be pruned without impacting the model quality.
Second, LLMs have much larger total parameter counts, which leads to even higher computation costs to apply first- and second-order method for pruning.

Most recently, Wanda~\cite{sun2023wanda} and SparseGPT~\cite{frantar2023sparsegpt} shows how to avoid the expensive computation to prune LLMs and successfully demonstrate a family of pruned LLM models.
OWL~\cite{yin2023outlier} improves on these techniques with an outlier-aware and non-uniform layer pruning. 
However, all of them can only prune and compress the model by about 2$\times$ without significant impacting the model quality. As we will see in later sections, compressing LLMs with pruning is still challenging. 

\subsection{Quantization}
Quantization is another popular method for DNN compression.
Contrary to sparsification, quantization aims to represent ``all" values in a tensor with a lower-bit-width format.
Since most models can be trained with \texttt{fp16} computation, deploying the trained model and performing inference with \texttt{fp16} is the natural extension.
As many researchers have found out~\cite{dai2021vsq}, once a model is trained, representing the weight value in lower-bit-width does not impact the model quality much.
For example, an original value in a higher precision can be rounded to the nearest (RTN) lower precision value as shown in \autoref{fig:sparse-quant}.
Besides the bit width of the quantization format, it is also crucial to decide the target for the quantization: weights and/or activations. 
When quantization is applied to both weights and activations, it reduces the cost of required hardware unit and enables the exploitation of simpler arithmetic computations with low-bit-width compared to the \texttt{fp16} computations, which saves area and power and increases compute throughput.
As a result, commercial products~\cite{nvidia_ampere, nvidia_hopper} have included \texttt{fp8} and \texttt{int8} computation support with a higher peak computation throughput for inference.

Similar to classic DNNs, LLMs can be quantized to lower bit precision such as \texttt{int8} or \texttt{fp8} after training. 
Previous work~\cite{llmint8_nipes22} shows how to quantize both weights and activations of LLMs with \texttt{int8} (W8A8) and leverage 8-bit matrix multiplication unit in modern hardware.
On the other hand, other work such as GPTQ~\cite{frantar2023optq} or AWQ~\cite{lin2023awq} focus on weight-only quantization to minimize the memory footprint of weights while keeping activations as is. 
Both GPTQ and AWQ show that it is able to quantize weights to 4 bits while maintaining the quality of LLMs (W4A16).
Although they reduce the bit width of weights aggressively, the weights are converted back to \texttt{fp16} and use lower-throughput \texttt{fp16} hardware units during the inference as activations are not quantized. 

An important problem in LLM quantization, as pointed out by recent work~\cite{llmint8_nipes22}, is that the quantization error becomes large due to a few \textbf{outliers} in activations, especially for the large models.
These outliers, by definition, are often a very small portion ($<10\%$) of the activation tensors.
Also, there are some \textbf{sensitive values} that would affect significantly to the final outcome.
As these values require to be stored and compute in a higher precision, computation savings with quantization is capped at 2$\times$ at most (\texttt{fp16} into \texttt{fp8}/\texttt{int8}).

\autoref{tab:previous_work} compares the mentioned prior work above, and our proposed method SDQ. Unlike the previous works, our method uses structured sparse decomposition with different bit widths to achieve 4$\times$ compute throughput. 

\section{Comparing Compression Methods}
\label{sec:metrics}
Since LLM inference stresses the system in various ways (computation, memory capacity, etc.), different compression methods often target different goals to optimize. 
In this section, we define a few metrics that can be used to estimate the computation and memory capacity savings for each method and highlight the critical design choices for them.

\subsection{Effective compute throughput for sparsification}
When multiplying a weight matrix with an activation matrix, if there is overall 90\% unstructured sparsity in weights, the required number of Multiply–Accumulate (MAC) operations can be reduced by 90\% by skipping ineffectual computations. 
However, identifying effectual computations and only mapping them to the compute unit (e.g., tensor cores) is not trivial, so it is difficult to exploit unstructured sparsity without investing significant area and power~\cite{qin2020sigma}.
To mitigate the problem, recent work from both industry and academia~\cite{zhu2019micro, nvidia_ampere, jeong:vegeta} propose to use \emph{N:M} structured sparsity (at most \emph{N} non-zero values in each consecutive \emph{M} values, as shown in \autoref{fig:sparse-quant}), which provides predictable compute throughput while exploiting sparsity for higher performance.
For example, with NVIDIA Ampere sparse tensor core, using 2:4 structured sparsity can reduce 50\% of the required number of MAC operations by skipping ineffectual computations. Using the 1:8 structured sparsity in an emerging Sparse Tensor Core~\cite{liu2021s2ta, jeong:vegeta} can reduce up to 87.5\%.
Generally, using \emph{N:M} structured sparse hardware support, one can reduce the required computation by $\frac{M-N}{M}\times100$\%) and increase the throughput respectively.

\textbf{In this work, we focus on using a futuristic, \emph{N:M} structured sparse tensor core support that can provide $\frac{M}{N}\times$ compute throughput for \emph{N:M} structured sparsity, which is more practical and realistic.}

\subsection{Effective compute throughput for quantization}
On the other hand, unlike sparsification, quantization methods are not directly used to reduce the number of MAC operations as there is no ineffectual computation after quantization.
However, if both weights and activations are quantized, the low-bit-width computation unit (e.g., INT8 Tensor Core) can be used, which is more efficient compared to the full-precision computation unit.
Thus, given the same area/power budget, the number of low-bit arithmetic units would be higher than that of high-bit arithmetic units, but the exact ratio could vary depending on the actual implementation. 
As mentioned earlier, the compute throughput of 4bit and 8bit format is 4$\times$ and 2$\times$ compared to that of the 16bit format, respectively in NVIDIA Ampere GPU~\cite{nvidia_ampere}. 

\textbf{In this work, we assume \texttt{fp16} as the baseline format for LLM inference, so we assume $\frac{16}{n}\times$ compute throughput can be achieved when both operands are quantized to $n$-bit for computations.}

\begin{figure}[!t]
	\centering
	\includegraphics[width=0.5\textwidth]{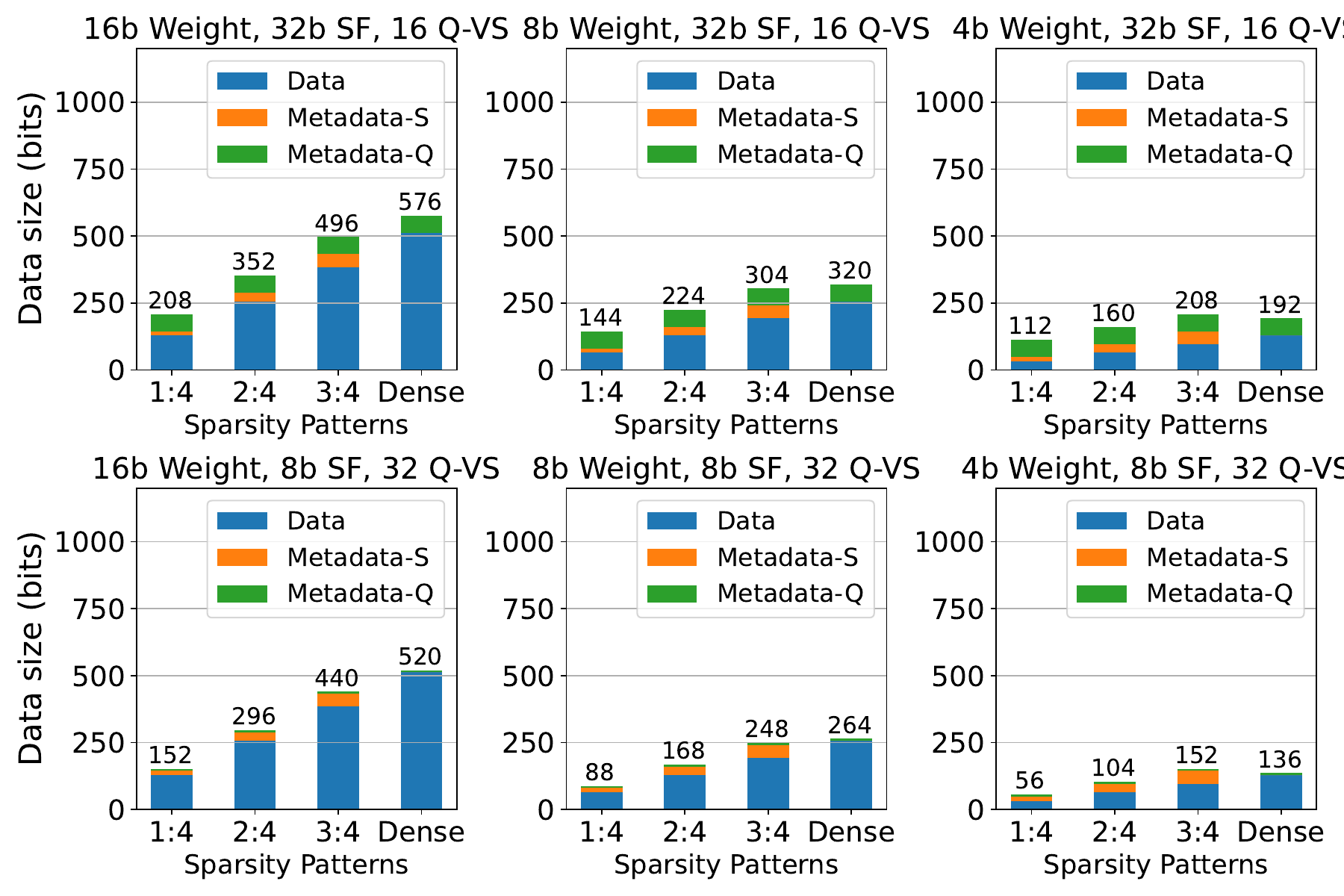}
\vspace{-1em}
	\caption{
 Data size for 32 elements with 1:4/2:4/3:4/Dense sparsity with quantization using 1) 32 bit scale factor with Q-Vector size of 16 (first row)
    2) 8 bit scale factor with Q-Vector size of 32 (second row). SF and Q-VS represent Scale Factor and  Q-Vector Size.}
\vspace{-1em}
	\label{fig:metadata}
\end{figure}

\subsection{Average bits per weight element}
The improved effective compute throughput mentioned above do not come for free.
To maximize the advantage of sparsification, weights are often stored in a compressed format, such as bitmask, ELLPACK, CSR, etc. 
A compressed format usually requires both data (for the actual non-zero values) and metadata (for the indexes of non-zero values).
Thus, the overhead in terms of memory is metadata to store the indexes of non-zero values. 
In \autoref{fig:sparse-quant}, we show how a simple vector composed of 8 values can be sparisfied to a vector with 2:4 structured sparsity. 
For \emph{N:M} sparsity, we call \emph{M} as the size of the S-Vector, so in this case, 4 is the size of the S-Vector. 
Both \emph{N} and \emph{M} could affect the size of the metadata.
For example, if we use ELLPACK-like format that stores indexes of non-zero elements in a vector, 2:4 sparsity requires 2 bits per index ($log_{2}4$) for each non-zero value, so $2\times2=4$ bits per vector, while 1:8 sparsity requires 3 bits per index ($log_{2}8$) for each non-zero value, so $3\times1=3$ bits per vector.

For quantization, the overhead that needs careful consideration is caused by scale factors.
In \autoref{fig:sparse-quant}, three different quantization method is used; round-to-neareast, round-to-nearest with a global scale factor, and round-to-nearest with per-vector scale factor (we call this as Q-Vector to distinguish with S-Vector).
A scale factor is used to match the range of quantized values and the range of the original values.
For example, without using a scale factor, all the values larger than 8 in \autoref{fig:sparse-quant} will be quantized to 8, which could cause a significantly large quantization error. 
As the finer granularity of the scale factor can provide more tuned quantization, it is better to have a smaller Q-Vector size for reducing quantization error.
However, similar to the indexes for compressed format, scale factors are the metadata for quantization. 
For example, assume a case where a value is quantized from 16-bit to 4-bit while a scale factor is 16-bit, and Q-Vector size is 4.
For each vector, the data size would be $4\times4=16$ bits while the metadata size is also 16 bits as the size of a scale factor is 16-bit (in this case, average bits per weight element would be $\frac{16+16}{4}=8$ bits).
Thus, scale factors are often also quantized and Q-Vector size is kept relatively large (16-64)~\cite{dai2021vsq}.

In \autoref{fig:metadata}, we show how much portion the size of metadata for sparsification (Metadata-S) and metadata for quantization (Metadata-Q) would take compared to the size of the actual data, showing that the metadata overhead caused by quantization and sparsification needs to be carefully considered as it could nullify the benefits of the optimization methods.
For example, a 3:4 sparse, 4-bit quantized model can have a higher bit-per-weight than a dense, 4-bit quantized model.


\section{SDQ: Tackling Challenges in LLM Compression with a Hybrid method}
\label{sec:motivation}
In~\autoref{tab:previous_work}, we summarize the previous work related to sparsification and quantization for LLMs.
The first observation from sparsification perspective is that even though 2:4/4:8 sparsity would provide 2$\times$ effective compute throughput, it is hard to use for LLMs~\cite{sun2023wanda, frantar2023sparsegpt} while maintaining the quality of the model (less than 1\% quality drop). 
We observed that even with the state-of-the-art sparsification methods, 2:4/4:8 sparsity on OPT-350M cause 77\%/142\% perplexity increase when evaluation with raw-Wikitext2 as shown in \autoref{tab:opt-perplexity}. Even though the larger models are more tolerant than small models, inducing 2:4/4:8 sparsity on the OPT-175B model (the largest public model) still causes 3$-$6\% perplexity increase, which is significant considering that only 1\% is acceptable by MLPerf benchmarking rules~\cite{mlperf}.
Also, even if 3$-$6\% perplexity increase is acceptable, it can only achieves 2$\times$ effective compute throughput.

The second observation is that the previous work focusing on weight-only quantization is able to reduce the weight bit width to 4 bit while dual quantization can only reduce the bit width to 8 bit~\cite{llmint8_nipes22} as it is harder to quantize both operands while maintaining the quality of the model.
Thus, even in the quantization perspective, it can only achieves 2$\times$ effective compute throughput using 8-bit computations leaving the opportunity of using 4-bit computations for 4$\times$ effective compute throughput due to the quality drop after quantization.

The last observation is that various recent work have noticed that isolating outliers is shown to be crucial to maintain quality after LLM quantization~\cite{llmint8_nipes22, dettmers2023spqr, lee2023owq, kim2023squeezellm}.
They have found that outliers in LLMs makes them hard to be quantized without causing huge error.
This is because even though there is a single outlier which is significantly larger than other values, that would make the corresponding scale factor much larger so that other small inliers would end up being quantized to the same value.

To resolve this issue, previous work extracted around 1-5\% outliers and stored using a higher precision separately in a compressed format.
During the inference, they treat these outliers as an unstructured sparse matrix and invoke SpMM kernels separately on either CPU or GPU CUDA Cores, instead of the Tensor Core.
This extra SpMM with unstructured sparse outliers is costly even with just 1\% outliers due to the huge compute throughput difference between CUDA Cores and Tensor Cores.
For example, in NVIDIA A100, the peak compute throughput difference between CUDA Cores and Tensor Cores is around 10$-$30$\times$ depending on the precision. 
We observed that even with 0.5\% sparsity ratio, using cuSparse for SpMM on CUDA cores is slower than treating the matrices as dense matrices and invoke a dense, accelerated GEMM on Tensor Cores~\cite{paulius_snn_talk}. 

Based on the three observations, we propose to extract local outliers with \emph{N:M} structrued sparse tensor instead of extracting global outliers with unstructured sparse tensor. 
As the previous work extract outliers globally, i.e., select elements in the entire tensor that has extreme values in a certain metric, the extracted global outliers are naturally unstructured sparse, which is not able to be accelerated through structured spare Tensor Core.
Instead, we propose to extract outlier per S-vector (we call this as local outliers), locally within a 1D vector. 
For example, assume that there is a vector composed of 16 elements with a single global outlier and we use 1:8 structured sparsity to extract local outliers for the given vector.
As we extract a local outlier per vector, a single global outlier would be guaranteed to be captured by 1:8 local outlier extraction.
If we assume there are two global outliers in the given vector, both global outliers can be extracted by 1:8 local outlier extraction only if they are in different S-vector.

It is possible that the two global outliers happen to be in the same S-vector, so in that case, the 1:8 local outlier extraction would not be able to capture that. 
For a rare case, when there are three global outliers in the given vector, the 1:8 local outlier extraction can miss 1-2 global outliers depending on the distribution of the outliers. 

Thus, \emph{N:M} local outlier extraction would only be effective if the outlier ratio is small and the location distribution is not extremely skewed. 
Fortunately, in LLMs, the percentage of outliers is reported around 1-5\%~\cite{guo2023olive, dettmers2023spqr}, which is small enough to use \emph{N:M} local outlier extraction without missing too most of the global outliers.
\begin{figure}[t]
	\centering
	\includegraphics[width=0.49\textwidth]{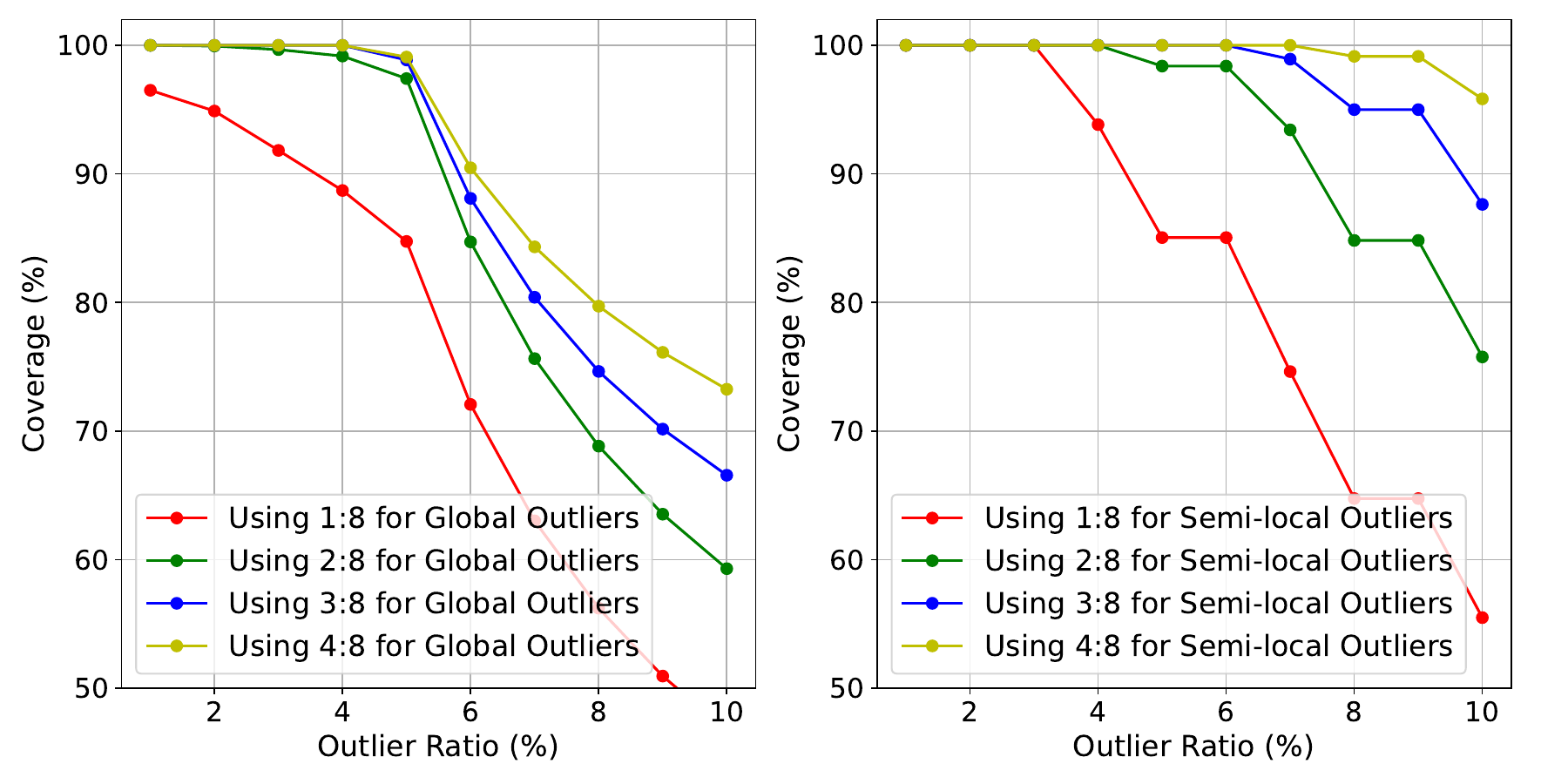}
\vspace{-1em}
        \caption{Local outlier extraction using \emph{N:8} for global outliers and semi-local outliers.}
\vspace{-1em}
    \label{fig:outlier_extraction}
\end{figure}
To demonstrate this, in the left plot of the \autoref{fig:outlier_extraction}, how much coverage \emph{N}:8 local outlier extraction show on one of the OPT-6.7B layer. 
As expected, we observe that 2:8 is enough capture 99\% of global outliers if the outlier ratio is smaller than 4\%. 

Moreover, \emph{N:M} local outlier extraction would be more efficient with the recent trend of finer granularity for the scale factors, i.e. smaller Q-Vector size.
As a separate scale factor would be used for each Q-Vector, the outlier extraction needs to capture semi-local (Q-Vector-wise) outliers, not the global outliers.

In the right plot of the ~\autoref{fig:outlier_extraction}, we show how \emph{N:M} local outlier extraction is able to cover semi-local outliers when the Q-Vector size is 64. 
As expected, \emph{N:M} local outlier extraction is able to show much higher coverage for semi-local outliers than global outliers, for example, 1:8 is enough to capture all semi-local outliers up to 3\% outlier ratio, thanks to its relatively regular outlier pattern.




\section{SDQ For LLMs}
\label{sec:sdq-sw}

In this section, we explain our framework for SDQ. As the name implies, it consists of three stages, sparsificaion, decomposition, and quantization, where each stage can leverage different existing methods and works together as a hybrid method.
\begin{figure}[t]
	\centering
	\includegraphics[width=0.49\textwidth]{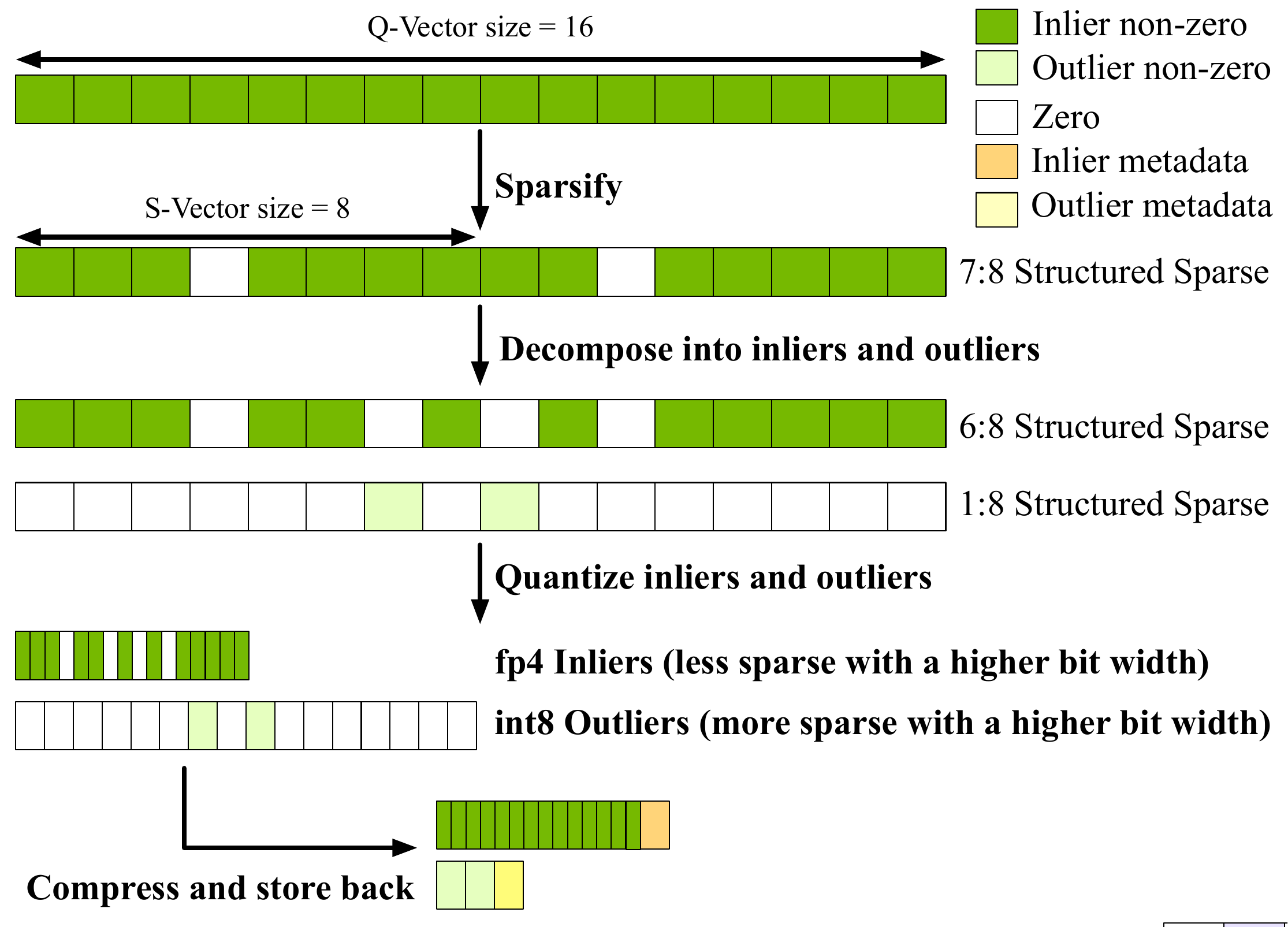}
\vspace{-1em}
        \caption{SDQ flow with a vector size of 16.}
\vspace{-1em}
	\label{fig:sdq_overview}
\end{figure}
\begin{figure*}[t]
	\centering
	\includegraphics[width=0.95\textwidth]{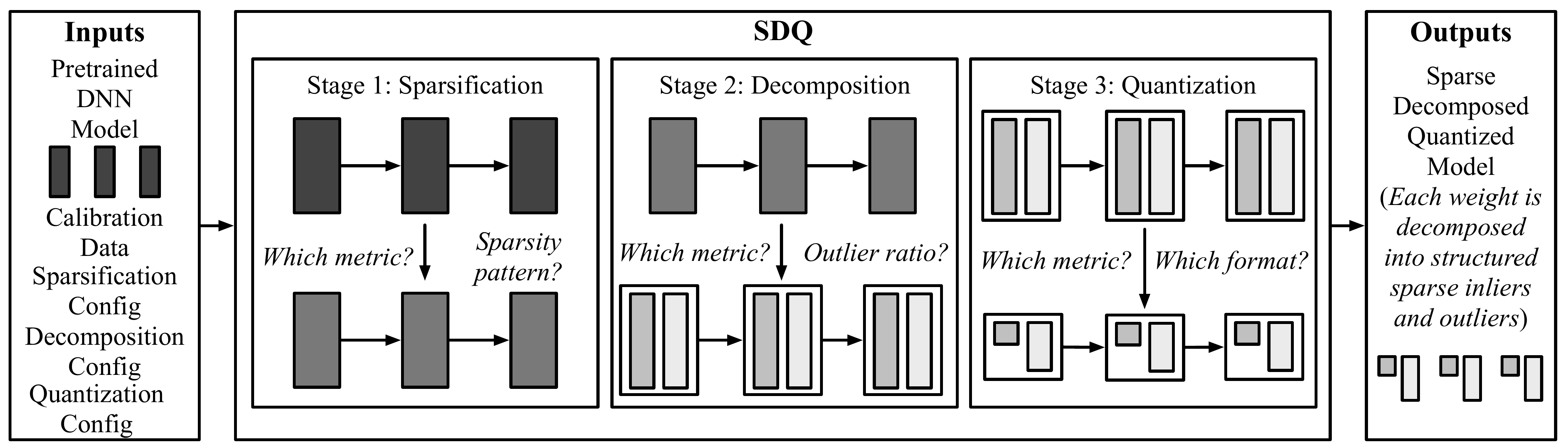}
\vspace{-1em}
        \caption{Overall SDQ framework composed of three stages.}
\vspace{-1em}
    \label{fig:framework}
\end{figure*}
\begin{figure}[!t]
	\centering
	\includegraphics[width=0.49\textwidth]{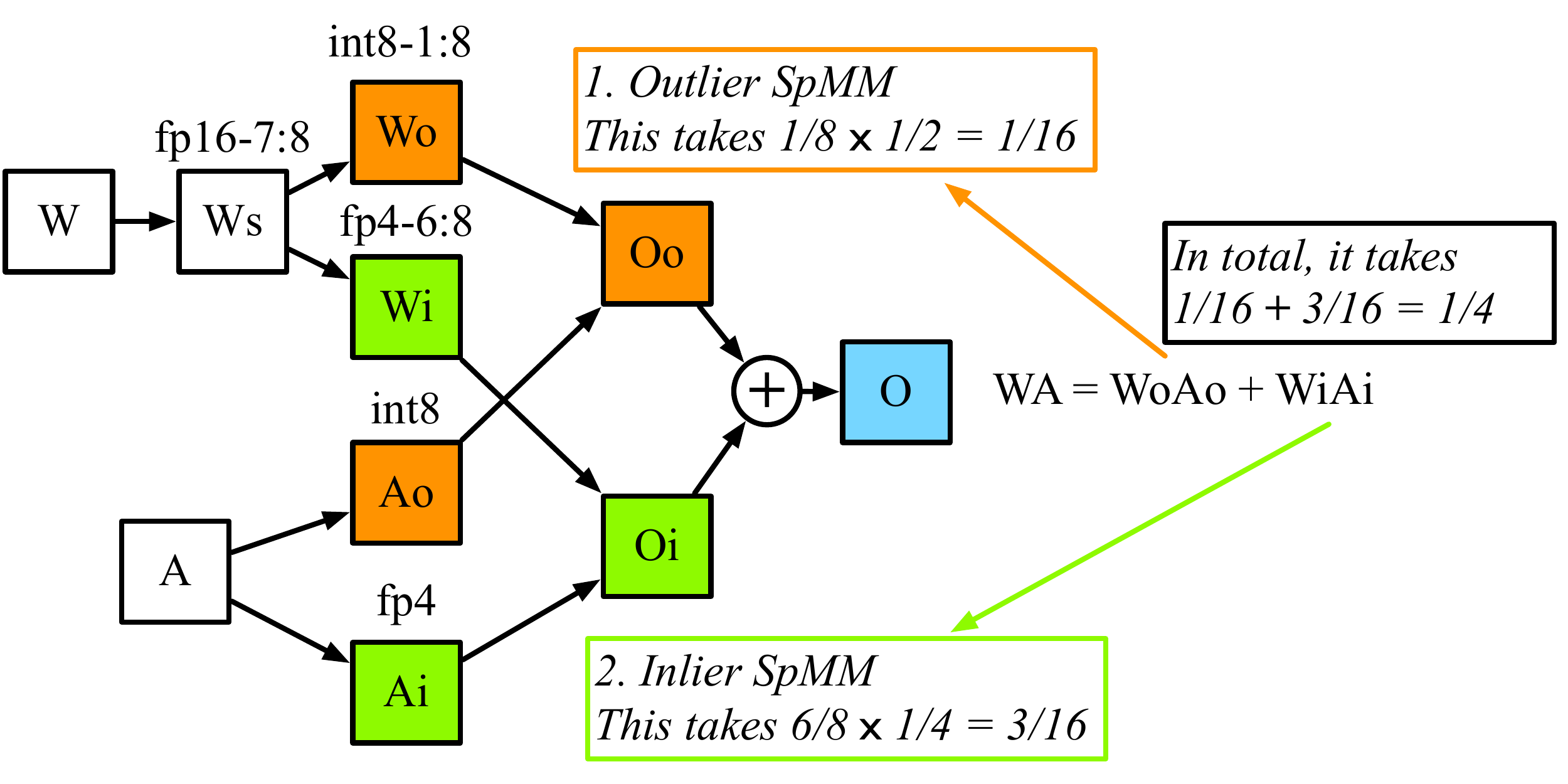}
        \vspace{-1em}
        \caption{Performance estimation with SDQ.}
\vspace{-1em}
    \label{fig:computation}
\end{figure}

\textbf{Stage 1 Sparsification}: SDQ first sparsify the weights of LLMs as much as possible, until the LLM quality are impacted signicantly (e.g., 1\% increase in perplexity).
SDQ prunes weights based on a significance metric under the structured pattern constraint, such as \emph{N:M}. 
The most straightforward metric is magnitude as the weight with a small magnitude is likely to not affect much to the final output~\cite{mishra2021accelerating}. 
The framework only keeps the largest \emph{N} values in each block that is composed of consecutive \emph{M} values.

If using calibration data is allowed, more sophisticated metrics could be used. 
For example, another metric is based on weight-activation-product (Wanda~\cite{sun2023wanda}) that is able to take the activation into account while choosing the significant weight values. 
Even though a weight value itself is tiny, it is possible that the corresponding activation is huge, increasing the impact to the output.
We also use SparseGPT~\cite{frantar2023sparsegpt} that is also using calibration data. Unlike the previous methods, SparseGPT updates weights in the sparsification to compensate for the error during the process and uses a Hessian-based metric as a significance metric.
In \autoref{fig:sdq_overview}, the sparsification stage uses 6:8 as the target structured sparsity using the given significance metric.
\begin{table*}[h]

\centering
\begin{tabular}{@{} cccccccc @{}}
\toprule
 & \multicolumn{5}{c}{OPT}                       \\ 
\cmidrule(l){2-8}
Optimization Configuration   & 125M   & 350M   & 1.3B  & 2.7B  & 6.7B  & 13B   & 30B \\ \hline
 \multicolumn{8}{c}{\textbf{1$\times$ Effective Compute Throughput}} \\
Dense-WA16   &  27.65 &  22.00 & 14.62 & 12.47 & 10.86 & 10.13 & 9.56   \\ 
 S-RTN-W4     &   -     & -       &   -    &  -     & 12.10 & 11.32 & 10.97     \\ 
 S-GPTQ-W4    &   -    &  -      &   -    &    -   & 11.39 & 10.31 & 9.63     \\ 
 S-SpQR-W4    &   -   &   -     &   -    &      - & 11.04 & 10.28 & 9.54     \\ 
 \hline 
 \multicolumn{8}{c}{\textbf{2$\times$ Effective Compute Throughput}} \\
 S-Wanda-4:8    & 53.12  & 58.90  & 22.21 & 16.77  &  13.55     & 13.35      &  10.87   \\ 
 S-SparseGPT-4:8&  43.99 & 38.92  &20.06  &15.04  &  12.57     &  11.83     & 10.31   \\ 
 Q-VSQuant-WA\texttt{int8}  & 27.77  & 22.01  & 14.75 & 12.50  & 10.95 & 10.14          &  9.57  \\ 
 Q-VSQuant-WAf\texttt{fp8}  & 27.78  & 22.05  & 14.75 & 12.51 & 10.95      & 10.13      &  9.58   \\ 
 \hline 
\hline 
 \multicolumn{8}{c}{\textbf{3.6$\times$ Effective Compute Throughput}} \\ 
  SDQ-8:8-1:8\texttt{int8}-7:8\texttt{fp4}   & 28.99  & 22.87  & 15.36  & 12.62  & 11.06      & 10.24    &  9.66 \\

 \hline 
  \multicolumn{8}{c}{\textbf{4$\times$ Effective Compute Throughput}} \\
 S-Wanda-2:8    &  1584.81& 2847.39  & 1217.73 &  6204.19 & 1028.50      & 2010.05      & 9997.60   \\ 
 S-SparseGPT-2:8&  849.99      & 645.33     & 990.81 & 150.09&  189.00     & 265.05      & 97.92     \\ 
 Q-VSQuant-WA\texttt{int4}  & 33.10  & 28.25  & 17.84 & 14.63  & 12.02 & 11.72          &  10.61  \\ 
 Q-VSQuant-WA\texttt{fp4}  & 30.50  & 24.31  & 16.00 & 13.29 & 11.32      & 10.57      &  9.90   \\  
 
 SDQ-W3:4-1:4\texttt{int8}-2:4\texttt{fp4}   & 31.42       & 26.76  & 16.33  & 13.11 & 11.17      & 10.41      &  9.63  \\ 
 SDQ-S3:4-1:4\texttt{int8}-2:4\texttt{fp4}   & 30.71      & 25.84  & 16.35  & 12.96  & 11.15      & 10.31       &  9.58  \\ 
 SDQ-W6:8-2:8\texttt{int8}-4:8\texttt{fp4}   & 30.15      & 24.48   & 15.71 &12.68  & 10.91      & 10.27 &  \textbf{9.51}   \\ 
 SDQ-S6:8-2:8\texttt{int8}-4:8\texttt{fp4}   & 29.70      & 24.45   & 16.41 & 12.58 & 10.98 & \textbf{10.20}      & 9.52   \\ 

 SDQ-W7:8-1:8\texttt{int8}-6:8\texttt{fp4}   & 29.30       & \textbf{23.02} & \textbf{15.44}  & 12.70  & \textbf{10.89}     &10.21       &  9.58  \\ 
 SDQ-S7:8-1:8\texttt{int8}-6:8\texttt{fp4}   & \textbf{28.82}     & 23.20 & 15.83  & \textbf{12.56}  & 10.98     &  10.26 &  9.60  \\ 
\bottomrule

\end{tabular}
\caption{Perplexity results of OPT on raw-WikiText2 with various sparsification, quantization, and SDQ configurations.}
\vspace{-1em}
\label{tab:opt-perplexity}
\end{table*}

\textbf{Stage 2 Decomposition}: In this stage, SDQ decomposes the the weight tensor of each layer into two tensors, inliers and outliers using the \emph{N:M} local outlier extraction explained in \autoref{sec:motivation}.
In \autoref{fig:sdq_overview}, it uses 1:8 as the target structured sparsity for the local outlier extraction so that the remaining values become the inliers.
An interesting point in this example after extracting outliers as 1:8 is that the remaining outlier becomes naturally structured sparse, in this case, 6:8. 

To choose outliers, we need another metric to decide whether a weight value is an outlier or not, similar to the significance metric for sparsification. 
We experiment with magnitude-based, activation-weight-product-based, and output-error-based metrics (based on the error after quantization). 
Unlike previous work, we make sure both inlier and outlier tensors are \emph{N:M} structured sparse so that we can utilize efficient SpMM on structured sparse HW.

\textbf{Stage 3 Quantization}: The last stage of SDQ is quantization where it uses different number formats for structured sparse inliers and outliers.
For quantization, we use VS-Quant~\cite{dai2021vsq} as it provides flexibility in the granularity of scale factors. 
We use a relatively higher bit width for outliers while using a lower bit width for inliers.
We also quantize activations accordingly so that we can use low-bit arithmetic computations that are cheaper in terms of area and power.
We show the high-level overview of the SDQ framework in \autoref{fig:framework}.
\subsection{LLM inference performance}
In \autoref{fig:computation}, we show how SDQ unveils 4$\times$ compute throughput.
As explained, the weights, $W$, is sparsified to $W_s$, then decomposed to $W_o$ and $W_i$. 
We quantize 1:8 $W_o$ as \texttt{int8} and 6:8 $W_i$ as \texttt{fp4}; accordingly, activation $A$ would be quantized to \texttt{int8} $A_o$ and \texttt{fp4} $A_i$.
Then, output activation from outliers $O_o$ is calculated through SpMM, which takes $\frac{1}{8}\times\frac{1}{2}=\frac{1}{16}$ using 1:8 and 8 bit computations compared to \texttt{fp16} baseline.
Similarly, output activation from inliers $O_i$ is computed through SpMM, which takes $\frac{6}{8}\times\frac{1}{4}=\frac{3}{16}$ using 6:8 and \texttt{fp4} computations compared to \texttt{fp16} baseline.
Thus, in total, it takes $\frac{1}{16}+\frac{3}{16}=\frac{1}{4}$, thus providing $4\times$ effective compute throughput.
The achieved effective compute throughput would depend on the target structured sparsity and number format, so we explore different configurations in \autoref{sec:eval}.

\section{Evaluation}
\label{sec:eval}

\begin{table*}[h]
\centering
\begin{tabular}{@{} cccccc @{}}
\toprule
 & \multicolumn{3}{c}{LLaMA-1} & \multicolumn{2}{c}{LLaMA-2}                       \\ 
\cmidrule(r){2-4}\cmidrule(l){5-6}
Optimization Configuration & 7B & 13B & 30B & 7B & 13B   \\ \hline      
\multicolumn{6}{c}{\textbf{1$\times$ Effective Compute Throughput}} \\
 Baseline                 & 5.68      & 5.09   & 4.10     & 5.12  & 4.57        \\ 
 S-RTN-W4                  & 6.43   & 5.55   & 4.57    & -      & -     \\  
 S-GPTQ-W4                 & 6.13   & 5.40   & 4.48     & -      & -     \\  
 S-SpQR-W4                 & 5.87   & 5.22   & 4.25      & -      & -     \\ 
 \hline
 \multicolumn{6}{c}{\textbf{2$\times$ Effective Compute Throughput}} \\  
 S-Wanda-4:8               & 8.57   & 7.41    & 5.97 &   8.07      &  6.55    \\ 
 S-SparseGPT-4:8           & 8.61   & 7.42     & 6.15 &  7.91      & 6.58   \\ 
 Q-VSQuant-WA-\texttt{int8}           & 5.70   & 5.11  & 4.13       & 5.14       & 4.60   \\ 
 Q-VSQuant-WA-\texttt{fp8}           & 5.70   & 5.11   & 4.12        & 5.14      & 4.59   \\  
 \hline
 \hline
 \multicolumn{6}{c}{\textbf{3.6$\times$ Effective Compute Throughput}} \\
 SDQ-8:8-1:8\texttt{int8}-7:8\texttt{fp4} & 5.81  & 5.20   & 4.22           & 5.23      & 4.67      \\ 

 \multicolumn{6}{c}{\textbf{4$\times$ Effective Compute Throughput}} \\\hline 
 S-Wanda-2:8      & 2960.99       & 2481.10 & 1228.86       & 1947.60      & 1188.45   \\ 
 S-SparseGPT-2:8  & 151.93       & 96.30     & 58.67       & 75.99      & 85.99    \\ 
 Q-VSQuant-WA-\texttt{int4} & 6.48       & 5.91   & 4.76                                   &   6.85      & 5.33   \\ 
 Q-VSQuant-WA-\texttt{fp4} & 5.97       & 5.29   & 4.33                                   &   5.36      & 4.74   \\ 
 SDQ-W3:4-1:4\texttt{int8}-2:4\texttt{fp4}           & 6.30    & 5.53  & 4.60                      &  5.67    &  5.00   \\ 
 SDQ-S3:4-1:4\texttt{int8}-2:4\texttt{fp4}           & 6.30    &  5.54     &   4.67                        &  5.65    &   4.98   \\ 
 SDQ-W6:8-2:8\texttt{int8}-4:8\texttt{fp4}  & 6.10   & 5.37  & 4.45                                 & 5.48     &  4.87  \\ 
 SDQ-S6:8-2:8\texttt{int8}-4:8\texttt{fp4}  &  6.10       &  5.36     &                 4.50                      &5.48               & 4.86       \\ 
 SDQ-W7:8-1:8\texttt{int8}-6:8\texttt{fp4}  & \textbf{5.87}   & \textbf{5.25}  & \textbf{4.27}      &    \textbf{5.30}  & \textbf{4.73}    \\ 
 SDQ-S7:8-1:8\texttt{int8}-6:8\texttt{fp4}  & 5.88                &  5.26             &      4.31              &    5.32   & \textbf{4.73}    \\ 

\bottomrule
\end{tabular}
\caption{Perplexity results of LLaMA-1/LLaMA-2 on raw-WikiText2 with various sparsification, quantization, and SDQ configurations.}
\vspace{-1em}
\label{tab:llama-perplexity}
\end{table*}

\subsection{Methodology}
In this section, we show how our hybrid method, SDQ, can be applied to LLMs to push the limit of effective compute throughput while achieving higher model quality than sparsification-only or quantization-only method.
We use OPT~\cite{zhang2022opt},  LLaMA~\cite{touvron2023llama}, LLaMA-2~\cite{touvron2023llama2} with different number of parameters, from 125M to 30B. 
We use the \texttt{fp16} dense version of each model as the baseline model.

First, we measure the perplexity change on the raw-Wikitext2 with OPT/LLaMA-1/LLaMA-2 from 125M to 30B parameters using various SDQ configurations.
Next, we provide zero-shot evaluation results of SDQ on OPT-6.7B, LLaMA-1-7B, and LLaMA-2-7B with BoolQ, HellaSwag, WinoGrande, ARC-e, ARC-c, and PIQA from LM-Eval~\cite{eval-harness}.
For baseline sparsification-only and quantization-only methods, we use SparseGPT~\cite{frantar2023sparsegpt}, Wanda~\cite{sun2023wanda}, and VS-Quant~\cite{dai2021vsq} with Q-Vector size of 16.
Only for the weight-only quantization methods (S-RTN-W4, S-GPTQ-W4, S-SpQR-W4), we use the results reported in the publication~\cite{dettmers2023spqr}.

We also compare different SDQ configurations. 
For example, SDQ-W7:8-1:8\texttt{int8}-6:8\texttt{fp4} means 1) using \textbf{W}anda (we use S for SparseGPT) as sparsification with \textbf{7:8} structured sparsity 2) and using \textbf{1:8} local outlier extraction with \texttt{int8} and \textbf{6:8} inlier with \texttt{fp4}.
For decomposition metric, we find the product-based one~\cite{sun2023wanda} performs the best, so we use the metric.
Starting with S- indicates sparsification-only and Q- indicates quantization-only (WA means dual quantization and W means weight-only).

\subsection{Perplexity evaluation}
We summarize the result of perplexity evaluation in \autoref{tab:opt-perplexity} for various OPT models.

\textbf{Comparison against sparsification.} 
As mentioned in ~\autoref{sec:motivation}, sparsification-only methods are not effective both for $2\times$ and $4\times$ effective compute throughput categories. 
For 2$\times$ category, S-SparseGPT-4:8 gives the best perplexity, increasing 8\% compared to the baseline model perplexity. 
Considering 1\% quality loss as the criteria, this is not acceptable; for instance, using baseline OPT-1.3B would be much better than using 4:8 OPT-2.7B, showing no benefits of using sparsification-only methods. 
This gets even worse for small models. For example, we found OPT-125M with the same configuration cause 59\% increase in terms of perplexity.
For 4$\times$ throughput category, the model gets totally broken with sparsification-only methods due to the not enough number of elements that the model can keep.  

\textbf{Comparison against weight-only quantization.} We observe that SDQ can achieve better perplexity than the best weight-only 4b quantization (\texttt{fp4}-e2m1) while enabling 4$\times$ effective compute throughput. 
For example, OPT-30B with SDQ-W6:8-2:8o8b-4:8i4b results in the perplexity of 9.51 which is better than that of S-SpQR-W4 (9.54) or even Dense-WA16 baseline (9.56).
\begin{table*}[t]
\centering
\begin{tabular}{@{} cccccccc @{}}
\toprule
\multicolumn{8}{c}{\textbf{OPT-6.7B}} \\ \hline
Method         & BoolQ     & HellaSwag & WinoGrande & ARC-easy & ARC-challenge  & PIQA & Average   \\ \hline      
Baseline     & 66.15  & 50.53     & 65.19      & 65.61 & 30.55  & 76.22    & 59.04  \\ 
\hline 
 S-SparseGPT-2:8  &  58.50         & 27.63      &  51.22    & 32.11    & 19.45  &56.20 & 40.85 \\ 
 S-Wanda-2:8      &  50.03      & 26.22       & 50.36     & 31.02& 18.69  & 55.33  & 38.61 \\ 
 Q-VSQuant-WA-\texttt{int4}  &   61.38           &  47.09      & 60.85         & 59.64      & 26.88     & 71.27 & 54.52\\ 
 Q-VSQuant-WA-\texttt{fp4}  & 62.94    & 48.50      & 61.80    & 65.15      & 29.18        & 74.70 & 57.05       \\ 
 SDQ-7:8-1:8\texttt{int8}-6:8\texttt{fp4} & \textbf{66.33}        & \textbf{49.91}      &  \textbf{63.54}     & \textbf{65.23} & \textbf{30.20}   & \textbf{74.97} & \textbf{58.36}  \\ 
\toprule
 \multicolumn{8}{c}{\textbf{LLaMA-1-7B}} \\ \hline
Baseline               & 75.11   & 56.95     & 69.85      & 75.29 & 41.89  & 78.67 & 66.29 \\ 
 \hline 
 S-SparseGPT-2:8         & 45.02   & 27.93     & 48.54     & 31.02 & 18.34  & 54.79 & 37.61    \\ 
 S-Wanda-2:8             & 37.83   & 26.35     & 50.04     & 26.52 & 20.65  & 53.43 & 35.80  \\ 
 Q-VSQuant-WA-\texttt{int4}         & 73.91  & 54.42     & 67.88     & 72.56 & 37.88  & 77.15  & 63.97  \\ 
 Q-VSQuant-WA-\texttt{fp4}        & 74.86    & 56.06     & \textbf{69.38}      & 74.54 & 40.87  & 77.26 & 65.49  \\ 
 SDQ-7:8-1:8\texttt{int8}-6:8\texttt{fp4}  & \textbf{75.35}    & \textbf{56.24}         & 69.30      & \textbf{74.70} & \textbf{41.38} & \textbf{78.29}     &  \textbf{65.88} \\ 
\toprule
 \multicolumn{8}{c}{\textbf{LLaMA-2-7B}} \\ \hline
Baseline            & 77.74      & 57.13     & 69.06      & 76.30 & 43.43  & 78.07 & 66.96 \\ 
\hline 
 S-SparseGPT-2:8  & 48.10         &  27.92     & 45.38     & 28.62 & 17.58 & 54.90 & 37.08  \\ 
 S-Wanda-2:8      & 37.83         & 26.15 & 50.91     & 26.94 & 19.62 & 52.82 & 35.71 \\  
 Q-VSQuant-WA-\texttt{int4} &   74.62         & 53.51 & 66.85     & 72.90 & 40.27     & 76.38 & 64.09 \\ 
 Q-VSQuant-WA-\texttt{fp4} & 75.81           & 55.99 & \textbf{67.96}      &  75.17 &41.38  & 77.26 & 65.60 \\ 
 SDQ-7:8-1:8\texttt{int8}-6:8\texttt{fp4}  & \textbf{78.44}   &    \textbf{56.43}       & 67.09            & \textbf{76.13} &  \textbf{43.17} & \textbf{77.58} & \textbf{66.47}  \\ 
\bottomrule
\end{tabular}
\caption{Zero-shot evaluation on various tasks with OPT-6.7B, LLaMA-1-7B, and LLaMA-2-7B.}
\vspace{-1em}
\label{tab:zeroshot}
\end{table*}

\textbf{Comparison against dual quantization.} 
Using dual quantization, one can achieve 2$\times$ and 4$\times$ if both weights and activations are quantized to 8-bit and 4-bit, respectively. 
For the 2$\times$ category, quantizing weights and activation with INT8 (or FP8) did not hurt the model quality, which is consistent to what the previous work~\cite{llmint8_nipes22} found.
For the 4$\times$ category, quantizing weights and activation with INT4 causes 9.9\%$-$19.7\% perplexity increases while that with FP4 causes 3.6\%$-$10.3\% perplexity increases, which is still significant quality loss based on the 1\% criteria.
Using SDQ, we observe that SDQ cause 0\%$-$4.2\% perplexity increaeses depending on the model sizes, which is significantly smaller than other 4$\times$ methods.
For the larger model, such as 2.7B, 6.7B, and 13B, we observe SDQ incurs less than 1\% perplexity increase, and interestingly, for OPT-30B, SDQ is able to reduce the perplexity, enabling 4$\times$ effective compute throughput with less than 1\% model quality drop.

We find the overall similar trend for LLaMA-1 and LLaMA-2 models as reported in \autoref{tab:llama-perplexity}.

\subsection{Zero-shot evaluation}
To provide a more holistic evaluation, we conduct experiments on various Zero-shot tasks and show the results in ~\autoref{tab:zeroshot}. 
These are known to provide more noisy results~\cite{llmint8_nipes22}, but still provide a big picture in terms of applicability.
As the individual result for each task could be noisy, we compare the average accuracy of all the zero-shot tasks.
To achieve 4$\times$ effective compute throughput, we observe the best sparsification-only and quantization-only methods show 25.58\% and 1.37\% accuracy drop, respectively, while SDQ only causes 0.53\% accuracy drop, showing the similar trend that we observe in the perplexity evaluation.
SDQ is the only option that meets the 1\% criteria.
\begin{figure}[!t]
	\centering
	\includegraphics[width=0.45\textwidth]{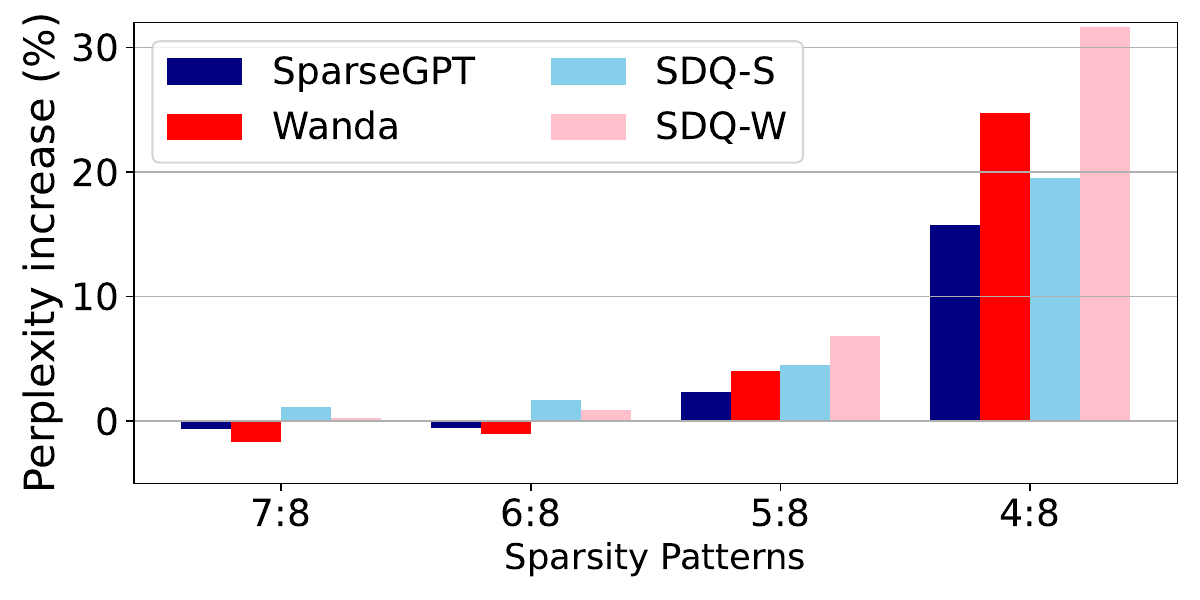}
        \vspace{-1em}
        \caption{Sensitivity study with different sparsification methods}
    \label{fig:sensitivity-sparsification}
    \vspace{-1em}
\end{figure}
\subsection{Sensitivity studies}
\label{subsec:sensitivity}
SDQ is composed of the three stages and the performance of the effectiveness of SDQ is also affected by the effectiveness of each stage. We use OPT-6.7B for this study.

\textbf{Sparsification stage.} In \autoref{fig:sensitivity-sparsification}, we show how different sparsity patterns and sparsification method affects to the effectiveness of SDQ. 
First, we observe that Wanda performs better than SparseGPT for 7:8 and 6:8 (even exceeding the baseline) while SparseGPT performs better than Wanda for 5:8 and 4:8.
Next, we apply SDQ with SparseGPT and Wanda for the sparsification stage with $N$:8 where $N$ is 7, 6, 5, and 4. 
We use 1:8 \texttt{int8} for outliers, and $N$-1:8 for inliers for each $N$ (as we use the same decomposition and quantization configuration for all SDQ options in this experiment, we just call them as SDQ-S and SDQ-W).
We observe the same trend for SDQ-S and SDQ-W, similar to the sparsification-only methods; for 7:8/6:8, SDQ-W exhibits lower (better) perplexity than SDQ-S while SDQ-S exhibits lower (better) perplexity than SDQ-W for 5:8/4:8.

\begin{figure}[!t]
	\centering
	\includegraphics[width=0.4\textwidth]{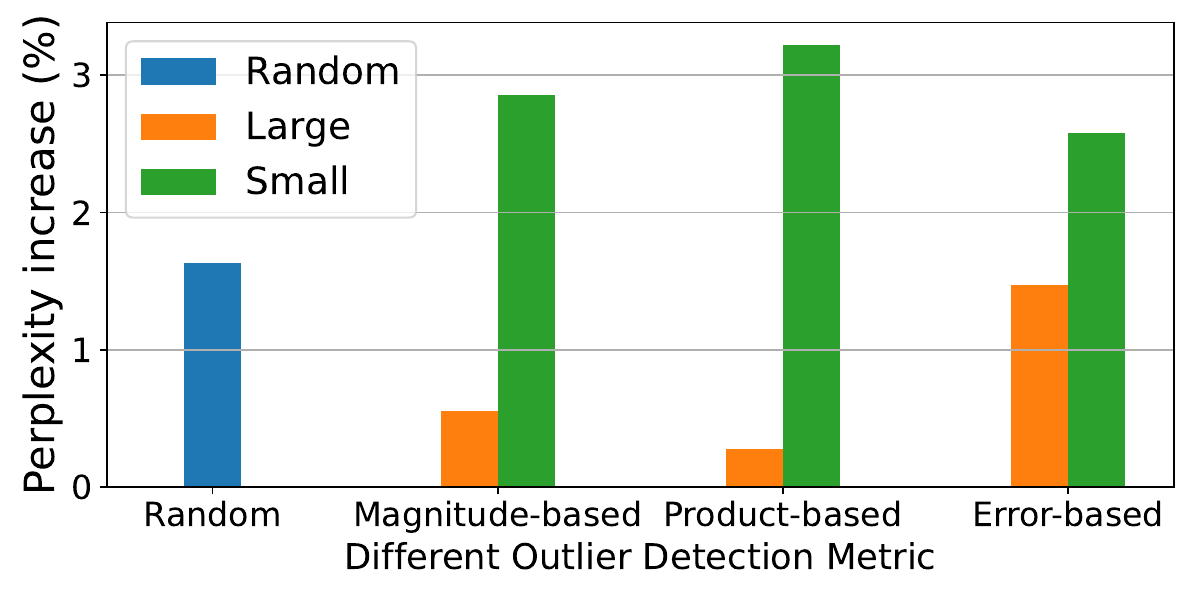}
        \vspace{-1em}
        \caption{Sensitivity study with different decomposition metrics.}
        \vspace{-1em}
    \label{fig:sensitivity-decomposition}
\end{figure}
\textbf{Decomposition stage.} In \autoref{fig:sensitivity-decomposition}, we show how different decomposition method affects to the effectiveness of SDQ.
We use SDQ-W7:8-1:8\texttt{int8}-6:8\texttt{fp4}, but with different criteria in the decomposition stage.
In the decomposition stage, a metric is required for \emph{N:M} local outlier extraction to identify the outliers. 
The simplest one is using the magnitude~\cite{guo2023olive}, but more sophisticated metrics could also be used such as weight-activation-product-based used in Wanda~\cite{sun2023wanda} or error-based similar to the one used in SpQR~\cite{dettmers2023spqr}.
Also, we can determine an element as an outlier based on the ascending or descending order of the metric. We mark ``Large" in the plot if we select outliers in descending order, and ``Small" if we select in ascending order.
We observe that product-based outlier extraction performs the best and the perplexity could fluctuate up to 7\% depending on the extracted outliers, implying that adopting an effective method to extract outliers is critical for the effectiveness of SDQ.
We leave exploring the best method to locate outliers for future work.
\begin{figure}[!t]
	\centering
	\includegraphics[width=0.4\textwidth]{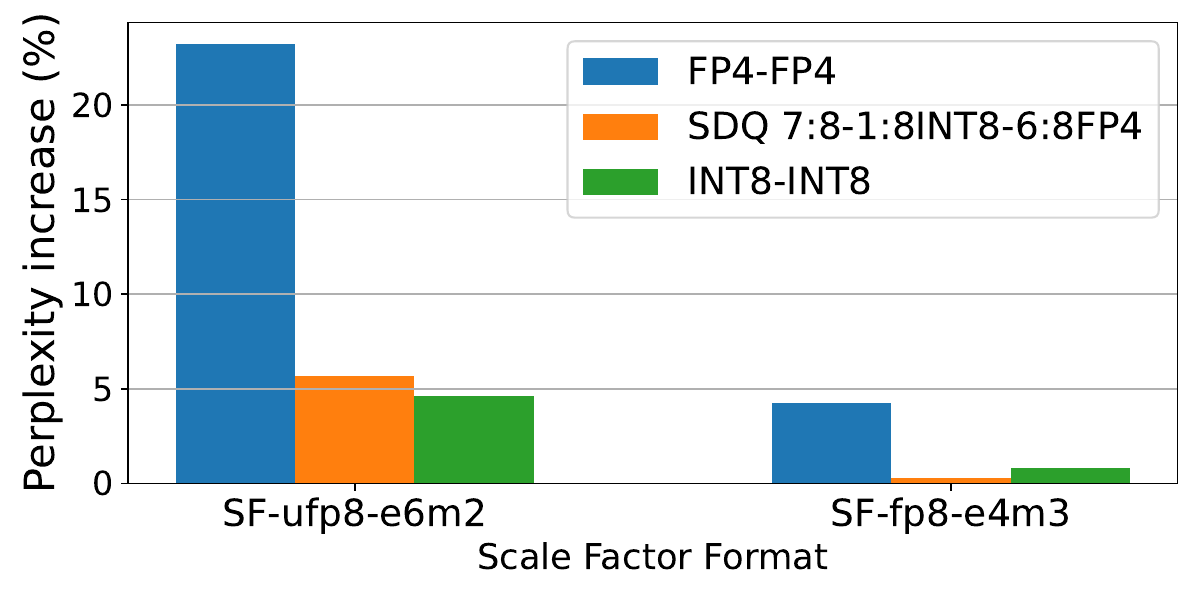}
        \vspace{-1em}
        \caption{Sensitivity study with different scale factor formats.}
\vspace{-1em}
    \label{fig:sensitivity-quantization}
\end{figure}
\textbf{Quantization stage.} VS-Quant can use different number formats for scale factor per Q-Vector.
In \autoref{fig:sensitivity-quantization}, we show how different quantization configurations, such as the scale factor format, affect the effectiveness of SDQ.
We compare two formats: \texttt{ufp8}-e6m2 (unsigned float using 6 bits for exponents and 2 bits for mantissa) and \texttt{fp8}-e4m3 (signed float using 4 bits for exponents and 3 bits for mantissa and 1 bit as the sign bit).
We observe that dual quantization using \texttt{fp4} with \texttt{ufp8}-e6m2 scale factors performs much worse than the case with \texttt{fp8}-e4m3. 
Similarly, dual quantization with \texttt{int8} also prefers \texttt{fp8}-e4m3 scale factors than \texttt{ufp8}-e6m2. 
We also observe that SDQ-W7:8-1:8\texttt{int8}-6:8\texttt{fp4} prefers \texttt{fp8}-e4m3, showing that improving quantization could also improve the quality of SDQ.

\section{Related Work}
\label{sec:related_work}
Both SparseGPT~\cite{frantar2023sparsegpt} and Wanda~\cite{sun2023wanda} propose sparsification methods for LLMs, but they both fail to maintain the quality of the model even with 50\% \emph{N:M} sparsity (such as 2:4 or 4:8).
Unlike sparsification, there have been many attempts to quantize LLMs~\cite{frantar2023optq, lin2023awq, dettmers2023spqr, llmint8_nipes22, xiao2023smoothquant, kim2023squeezellm}.
We use VS-Quant~\cite{dai2021vsq} in the current SDQ, but our work is not limited to the quantization method as shown in \autoref{subsec:sensitivity}. 
We believe SDQ can take the benefits of the improvement in quantization methods.

OBC~\cite{frantar2022obc} provides a framework for quantization and sparsification, but they do not target LLMs. Also, they do not consider running LLMs efficiently through \emph{N:M} structured sparse HW and low-bit computations.
Recent work~\cite{kuzmin2023pruning} observe that quantization generally performs better than sparsification on Vision models, which is consistent with our observation for LLMs.

\section{Conclusion}
\label{sec:conclusion}

We propose a new method, Sparse Decomposed Quantization (SDQ) using mixed precision for outliers and inliers through structured decomposition, which enables utilizing mixed precision \emph{N:M} structured sparse HW.
We explore the SDQ across different models and sizes to understand the potential of the technique with various LLMs.
We show that SDQ opens up an opportunity to achieve 4$\times$ effective compute throughput while maintaining the model quality ($<1\%$ quality drop). We plan to extend this to validate with sparse tensor accelerator simulators, such as Sparseloop~\cite{wu2022sparseloop}.

\newpage

\bibliography{example_paper}
\bibliographystyle{mlsys2024}


%


\end{document}